\documentclass{article}
\usepackage{spconf,amsmath,graphicx}

\usepackage{amsmath}
\usepackage{comment}
\usepackage{amssymb}
\usepackage{svg}
\usepackage{enumitem}

\usepackage{hyperref}
\setlist[itemize]{label=\textbullet}
\usepackage{multirow}
\usepackage[normalem]{ulem}
\useunder{\uline}{\ul}{}
\usepackage{booktabs}
\usepackage{float}

\usepackage[skip=5pt]{caption} 
\usepackage{setspace}
\usepackage{wrapfig}

\title{BIG-MoE: Bypass Isolated Gating MoE for Generalized Multimodal Face Anti-Spoofing}
\name{Yingjie Ma$^{1,2}$, Zitong Yu$^{2,3 *}$, Xun Lin$^{2}$, Weicheng Xie$^{1,4}$, Linlin Shen$^{1,3,4 *}$\thanks{* Corresponding authors}} 
\address{
    $^{1}$College of Computer Science and Software Engineering, Shenzhen University \ \ 
    $^{2}$Great Bay University \\
    $^{3}$National Engineering Laboratory for Big Data System Computing Technology, Shenzhen University \\
    $^{4}$Guangdong Provincial Key Laboratory of Intelligent Information Processing  \ \
}
\begin{document}

\maketitle

\begin{abstract}
In the domain of facial recognition security, multimodal Face Anti-Spoofing (FAS) is essential for countering presentation attacks. However, existing technologies encounter challenges due to modality biases and imbalances, as well as domain shifts. Our research introduces a Mixture of Experts (MoE) model to address these issues effectively. We identified three limitations in traditional MoE approaches to multimodal FAS: (1) Coarse-grained experts' inability to capture nuanced spoofing indicators; (2) Gated networks' susceptibility to input noise affecting decision-making; (3) MoE's sensitivity to prompt tokens leading to overfitting with conventional learning methods. To mitigate these, we propose the Bypass Isolated Gating MoE (BIG-MoE) framework, featuring: (1) Fine-grained experts for enhanced detection of subtle spoofing cues; (2) An isolation gating mechanism to counteract input noise; (3) A novel differential convolutional prompt bypass enriching the gating network with critical local features, thereby improving perceptual capabilities. Extensive experiments on four benchmark datasets demonstrate significant generalization performance improvement in multimodal FAS task. The code is released at \textcolor{purple}{\href{https://github.com/murInJ/BIG-MoE}{https://github.com/murInJ/BIG-MoE}}.
\end{abstract}


\vspace{-0.8em}
\section{Introduction}
\vspace{-0.8em}
Face Recognition (FR) technology, celebrated for its efficiency and accuracy in applications such as security surveillance and mobile payments, now confronts escalating security threats from sophisticated face rendering attacks. Traditional FR systems struggle to discern these attacks, which include printed photos, video playback, and 3D masks, underscoring the urgent need for robust security measures.

\begin{figure}[t]
\vspace{-1.8em}
\centering
\includegraphics[width=0.45\textwidth]{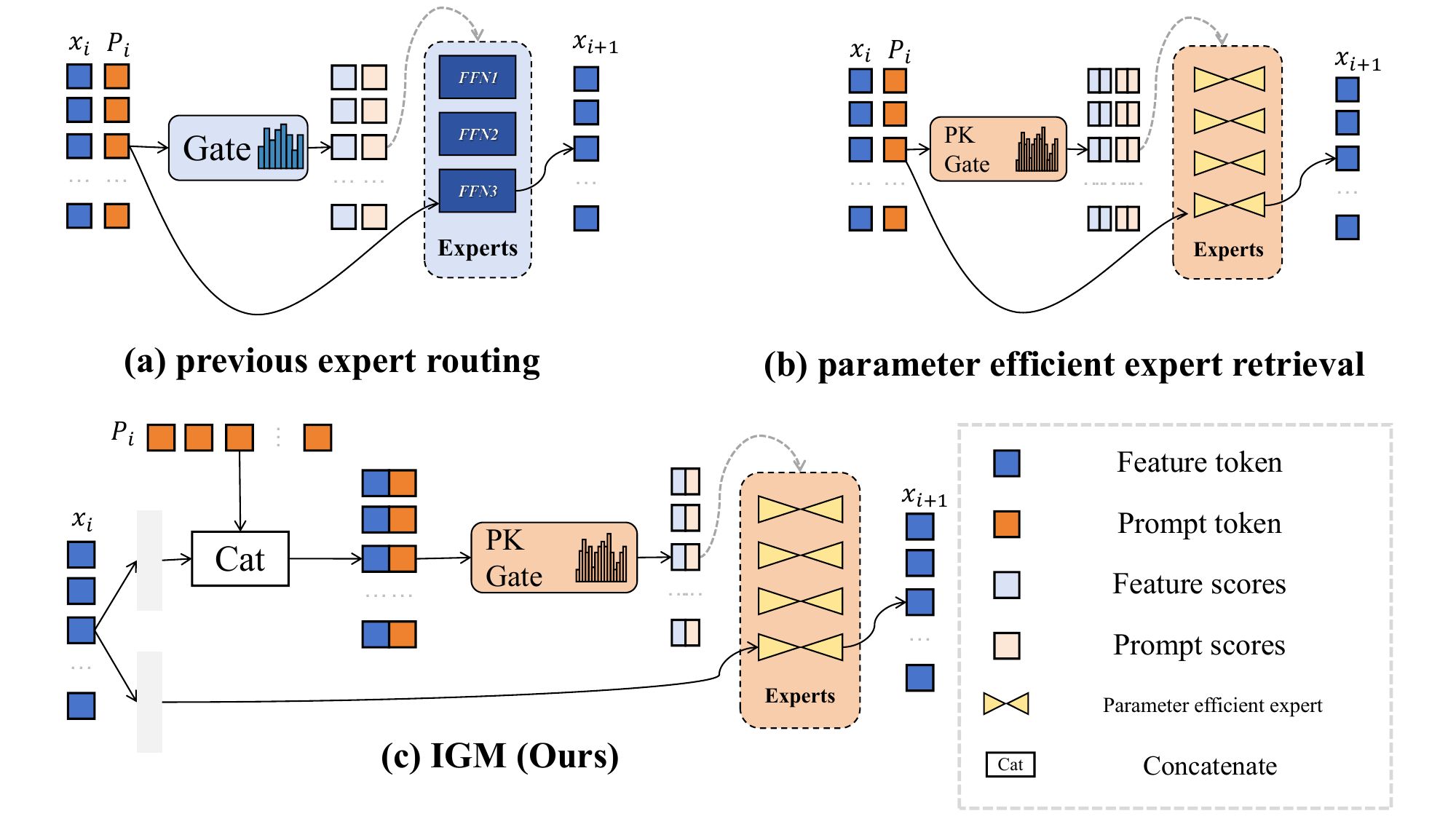}
 \vspace{-0.8em}
\caption{\textbf{Existing MoE prompt learning paradigm vs. Ours.} Both (a) conventional MoE prompt learning and (b) the parameter-efficient expert retrieval \cite{he2024mixture} approaches input prompt and feature tokens into the gating network or product key gating (PK Gate) network to generate scores for expert selection and subsequent processing with different gating mechanisms and types of experts. (c) Our Isolated Gating Mechanism (IGM) concatenates prompt and feature tokens for gating network scoring, and then processes feature tokens exclusively, isolating expert network input to enhance noise resilience and processing precision.}
\label{IGM}
 \vspace{-0.8em}
\end{figure}

\begin{figure*}[t]
\vspace{-2.8em}
\centering
\includegraphics[width=0.8\textwidth]{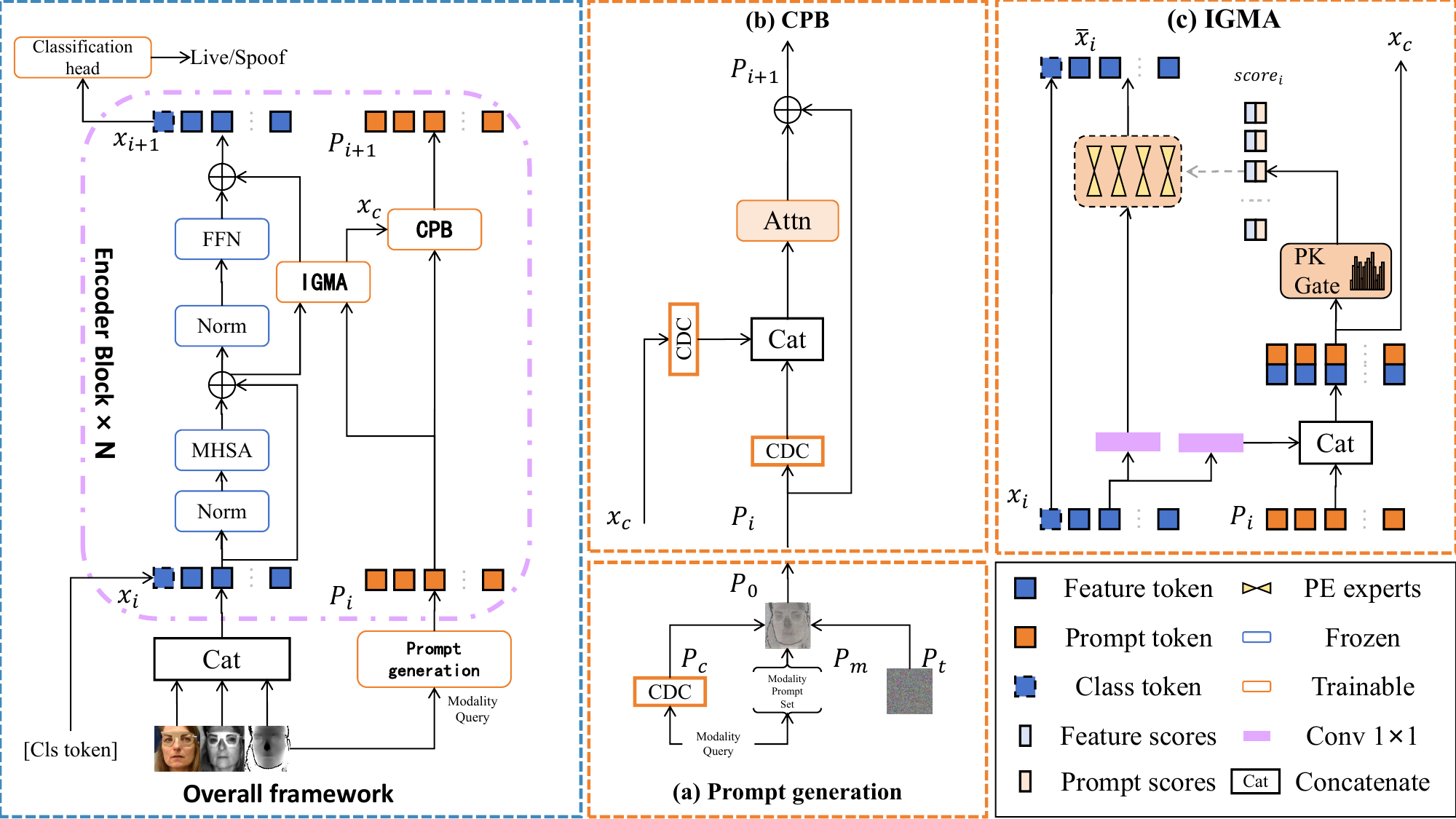}
\vspace{-0.3em}
\caption{\textbf{BIG-MoE Framework Overview}: The diagram succinctly captures the essential process and components of our approach: (a) \textbf{Prompt Generation}: This step outlines the creation and integration of initial prompts. (b) \textbf{CPB}: Describes the Convolutional Prompt Bypass, focusing on its enhancement of feature extraction via Central Difference Convolution (CDC) \cite{yu2020multi} and multimodal prompt integration. (c) \textbf{IGMA}: Highlights the Isolated Gating Mechanism Adapter's role in gating and its interaction with CPB across layers, promoting information exchange for enhanced model performance and robustness.}
\label{model}
\vspace{-1.5em}
\end{figure*}

To counter these threats, the research community has turned to Face Anti-Spoofing (FAS) techniques, which differentiate between genuine and spoofed faces \cite{yu2022deep}. Multimodal FAS methods \cite{george2021cross,george2020learning,liu2023ma,liu2023fm,yu2020multi,yu2023visual,yu2024rethinking}, integrating information from RGB images, depth maps, and infrared images, have shown promise in capturing comprehensive physical and behavioral facial features \cite{yu2023flexible}. However, the integration of multimodal data is challenging, often hampered by inter-modal feature bias and imbalance, and subtle multimodal spoof clues are easily drowned by domain shifts caused by sensor/environment discrepancy \cite{zhang2023provable,lin2024suppress}.



The Mixture of Experts (MoE) model, adept at handling complex data distributions, decomposes a large network into specialized smaller networks, reducing computational load through sparse activation and enhancing model generalization \cite{riquelme2021scaling,fedus2022switch,zoph2022st,xue2024openmoe}. This architecture excels in multi-task and multi-modal learning scenarios, especially with high dimensional and heterogeneous data \cite{yang2024multi}. MoE has also shown excellent results for sparse representations in FAS tasks \cite{zhou2022adaptive,kong2024moe,liu2024moeit}. Building on this, our research integrates fine-grained experts \cite{he2024mixture} into the MoE framework for multimodal FAS tasks, improving the capture of detailed data features crucial for FAS performance \cite{zoph2022st,xue2024openmoe}. To counteract the vulnerability to input noise, we propose an Isolation Gating Mechanism, depicted in Fig. \ref{IGM}, which processes input vectors to robustly understand feature relationships without direct fitting.

Furthermore, we explore the integration of prompt learning \cite{zhu2023visual} within the MoE model. To maximize prompt learning's benefits while avoiding feature confusion, we introduce a Convolutional Prompt Bypass (CPB) that interacts with the gated network to capture local deception features without direct feature propagation. Our contributions are threefold:
\begin{itemize}
\item We proposed the BIG-MoE, a novel multimodal FAS architecture that pioneers the application of MoE with fine-grained experts. This pioneering approach allows for more effective extraction of subtle cues and integration of multimodal features.
\vspace{-0.5em}
\item The BIG-MoE framework features an Isolated Gating Mechanism to shield the model against input noise and includes a convolutional prompt bypass, which fortifies the gating network with essential cues, thereby enhancing the model's robustness against overfitting and noise.
\vspace{-1.7em}
\item Extensive experiments demonstrate the reliability and superior performance of BIG-MoE for generalized multimodal FAS.
\vspace{-0.8em}
\end{itemize}

\begin{table*}[t]
\centering

\vspace{-1.2em}
\caption{Cross-dataset testing results under the fixed-modal scenarios (Protocol 1) among CASIA-CeFA (C) \cite{liu2021casia}, PADISI (P) \cite{rostami2021detection}, CASIA-SURF (S) \cite{zhang2020casia}, and WMCA (W) \cite{george2019biometric}. DG, MM, and FM are short for domain-generalized, multi-modal, and flexible-modal, respectively. Best results are marked in \textbf{bold}.
} 
\vspace{-0.5em}
\resizebox{0.99\textwidth}{!}{\begin{tabular}{c|c|cc|cc|cc|cc} 
\hline
  \multirow{2}{*}{\textbf{Method}}&\multirow{2}{*}{\textbf{Type}}& \multicolumn{2}{c|}{\textbf{CPS$\rightarrow$W}}&\multicolumn{2}{c|}{\textbf{CPW$\rightarrow$S}}&\multicolumn{2}{c|}{\textbf{CSW$\rightarrow$P}}&\multicolumn{2}{c}{\textbf{PSW$\rightarrow$C}} \\
  
  & & HTER(\%)$\downarrow$ & AUC(\%)$\uparrow$ & HTER(\%)$\downarrow$ & AUC(\%)$\uparrow$& HTER(\%)$\downarrow$ & AUC(\%)$\uparrow$ & HTER(\%)$\downarrow$ & AUC(\%)$\uparrow$ \\
  \hline
  SSDG \cite{jia2020single} (ECCV2022)&DG&26.09&82.03&28.50&75.91&41.82&60.56&40.48&62.31\\
  SSAN \cite{wang2022domain} (CVPR2022)&DG&17.73&91.69&27.94&79.04&34.49&68.85&36.43&69.29\\
  IADG \cite{zhou2023instance} (CVPR2023)&DG&27.02&86.50&23.04&83.11&32.06&73.83&39.24&63.68\\
  ViTAF \cite{huang2022adaptive} (ECCV2022)&DG&20.58&85.82&29.16&77.80&30.75&73.03&39.75&63.44\\
  MM-CDCN \cite{yu2020multi} (CVPR2020)&MM&38.92&65.39&42.93&59.79&41.38&61.51&48.14&53.71\\
  CMFL \cite{george2021cross} (CVPR2021)&MM&18.22&88.82&31.20&75.66&26.68&80.85&36.93&66.82\\
  ViT+AMA \cite{yu2024rethinking} (IJCV2024)&FM&17.56&88.74&27.50&80.00&21.18&85.51&47.48&55.56\\
  VP-FAS \cite{yu2023visual} (arXiv 2023) &FM&16.26&91.22&24.42&81.07&21.76&85.46&39.35&66.55\\
  MMDG \cite{lin2024suppress} (CVPR2024)&MM&\textbf{12.79}&\textbf{93.83}&15.32&92.86&\textbf{18.95}&\textbf{88.64}&29.93&76.52\\
  \hline
  ViT \cite{dosovitskiy2020image}&Baseline&20.88&84.77&44.05&57.94&33.58&71.80&42.15&56.45\\
  \textbf{BIG-MoE}&Ours&17.4&90.87&\textbf{10.96}&\textbf{94.35}&19.62&84.72&\textbf{7.71}&\textbf{97.72}\\
  \hline
\end{tabular}}

\label{res1}
\vspace{-1.3em}

\end{table*}

\vspace{-0.8em}
\section{Methodology}
\vspace{-0.8em}

\begin{table}[t]
\centering
\caption{Cross-dataset testing results under the limited source domain scenarios (Protocol 3) among CeFA-CeFA (C) \cite{liu2021casia}, PADISI USC (P) \cite{rostami2021detection}, CASIA-SURF (S) \cite{zhang2020casia}, and WMCA (W) \cite{george2019biometric}. Best results are marked in \textbf{bold}.
} 
\vspace{-0.5em}
\resizebox{0.49\textwidth}{!}{\begin{tabular}{c|cc|cc} 
\hline
  \multirow{2}{*}{\textbf{Method}}  & \multicolumn{2}{c|}{\textbf{CW$\rightarrow$PS}}&\multicolumn{2}{c}{\textbf{PS$\rightarrow$CW}} \\
  & HTER(\%)$\downarrow$ & AUC(\%)$\uparrow$ & HTER(\%)$\downarrow$ & AUC(\%)$\uparrow$\\
  \hline
  SSDG \cite{jia2020single} (ECCV2022)&25.34&80.17&46.98&54.29\\
  SSAN \cite{wang2022domain} (CVPR2022)&26.55&80.06&39.10&67.19\\
  IADG \cite{zhou2023instance} (CVPR2023)&22.82&83.85&39.70&63.46\\
  ViTAF \cite{huang2022adaptive} (ECCV2022)&29.64&77.36&39.93&61.31\\
  MM-CDCN \cite{yu2020multi} (CVPR2020)&29.28&76.88&47.00&51.94\\
  CMFL \cite{george2021cross} (CVPR2021)&31.86&72.75&39.43&63.17\\
  ViT+AMA \cite{yu2024rethinking} (IJCV2024)&29.25&76.89&38.06&67.64\\
  VP-FAS \cite{yu2023visual} (arXiv 2023) &25.90&81.79&44.37&60.83\\
  MMDG \cite{lin2024suppress} (CVPR2024)&\textbf{20.12}&\textbf{88.24}&36.60&70.35\\
  \hline
  ViT \cite{dosovitskiy2020image} (Baseline)&42.66&57.80&42.75&60.41\\
  \textbf{BIG-MoE (Ours)}&22.35&83.50&\textbf{14.11}&\textbf{95.13}\\
  
  \hline
\end{tabular}}
\label{limited}
 \vspace{-1.0em}
\end{table}

As shown in Fig. \ref{model}, our proposed Bypass Isolated Gating MoE (BIG-MoE) framework is fundamentally composed of a pre-trained Vision Transformer (ViT), coupled with a sophisticated prompt generation module, the Convolutional Prompt Bypass (CPB), and the Isolated Gating Mechanism Adapter (IGMA). Input data is transformed into visual prompt tokens by the prompt generation module, which are then enhanced by the CPB module. Concurrently, the input is processed through the ViT Encoder and IGMA, with the latter leveraging the CPB's visual prompts to augment gating perception. The aggregated outputs from both modules are fed into a classifier, and the predictions are refined by cross-entropy loss during backpropagation.

\vspace{-0.8em}
\subsection{Isolated Gating Mechanism Adapter}
\vspace{-0.3em}
Traditional MoE architectures are constrained by routing overhead in fine-grained expert partitioning. The PEER \cite{he2024mixture} architecture, however, employs the Product Key Retrieval (PKR) technique to efficiently identify and retrieve top-$k$ experts from a large pool for a $d$-dimensional input vector $x$, using low-dimensional sub-keys to construct a key set and inner product calculations, thus reducing computational load while preserving accuracy. The gating network, parameterized by $\Theta$, refines the expert outputs, incorporating a noise term $R_{noise}$, to yield the final gating decision $G(x;\Theta)$.

The sensitivity of the gating network to input noise escalates with an increasing number of fine-grained experts, which, despite the introduction of training noise, fails to optimize performance or fully exploit multimodal processing. We attribute this to the gating network's constrained feature perception due to low-dimensional sub-keys, impacting noise robustness. To address this, we introduce an IGM that distinguishes the expert-processed vector $x_e$ from the gating vector $x_g$, enabling a more nuanced nonlinear transformation to reduce noise impact and enhance system performance efficiently. This refined process is formalized as follows:
\begin{equation}
    F_{\text{MoE}}(x;\Theta,\{W_i\}^N_{i=1})=\sum^N_{i=1}G(x;\Theta)_iF_i(x;W_i)
\end{equation}
\begin{equation}
   G(x;\Theta)=\text{softmax}(\text{TopK}(f_g(\text{PKR}(x;\Theta)+R_{\text{noise}},k)))_i
\end{equation}
\begin{equation}
    F(x;\Theta)=f(f_e(x;W_e),W)
\end{equation}

The refined formulation shows that the input vector $x$ is processed by $f_g$ for gating and retrieval, followed by further processing of $f_e$ based on gating results to produce the final output $F(x;\Theta)$. This approach integrates fine-grained experts with IGM, resulting in the IGMA, as illustrated in Fig. \ref{model}(c).

\vspace{-0.8em}
\subsection{Convolutional Prompt Bypass}
\vspace{-0.8em}
Previous research has shown that routing selection in MoE models is sensitive to prompt tokens \cite{xue2024openmoe}, which can introduce noise and limit the effectiveness of Prompt Learning when applied to MoE. To address this, we developed the CPB for the IGMA, utilizing Central Difference Convolution (CDC) \cite{yu2020multi} to enhance the extraction of local spoofing cues.

The CPB process initiates by concatenating multimodal inputs along the channel dimension to create clue prompts $P_c$. A 30\% probability masks entire modal images, setting them to zero, which is integrated into the prompts $P_m$ as supplemental data. Static task-related prompts $P_t$ are concurrently acquired. These prompts are merged to form a comprehensive input prompt $P$. Each layer's prompt $P_i$ is combined with the perceptive vector $x_g$, forming an integrated perceptive vector input to the gating network. This fusion enhances perceptual stability, particularly with composite features exhibiting substantial representational variance.

The PKR method is employed to partition the perceptive vector into two sub-spaces, avoiding interference from prompt semantics and enhancing perception stability. The combined perceptive vector $ x_g $ is processed through the Attention mechanism, generating a new prompt $ P_{i+1} $ for the next layer, as described by the formula:
\begin{equation}
    P_{i+1} = P_i + \text{Attn}(\text{Cat}(P_i, x_c))
\end{equation}

Here, the Efficient Channel Attention (ECA) module within the CPB enriches IGMA with supplemental perceptive information, blending insights across layers to reduce gating sensitivity and bolster the model's performance and stability.

\vspace{-0.8em}
\section{Experiment}
\vspace{-0.8em}
\subsection{Data and Evaluation Metrics}
\vspace{-0.3em}
In this study, we followed the MMDG's Protocols 1 and 3 \cite{lin2024suppress}, applying a Leave-One-Out (LOO) test on fixed modalities: S (SURF) \cite{zhang2020casia}, P (PADISI USC) \cite{rostami2021detection}, C (CeFA) \cite{liu2021casia}, and W (WMCA) \cite{george2019biometric}. Performance was measured using Half Total Error Rate (HTER) and Area Under the Receiver Operating Characteristic Curve (AUC).

\vspace{-1.0em}
\subsection{Implementation Details}
\vspace{-0.8em}
All input images were standardized to 224 × 224 × 3 pixels, segmented into 14 × 14 patches, and inputted into the ViT where token hidden dimension $d$=768. We trained the model using the Adam optimizer, a learning rate of 5e-5, weight decay of 1e-3, over 100 epochs with a batch size of 32. The classifier was a single fully connected layer reducing the class token output from 768 to 2. The model was based on a pre-trained ViT-Base on ImageNet, with the IGMA structure featuring 2 activated experts per head, in total 1600 experts, a hidden dimension of 8, and a 64-dimension CPB. 

\vspace{-1.0em}
\subsection{Cross-testing Results}
\vspace{-0.5em}
\noindent\textbf{Sufficient Source Domains Scenario.} \quad   
The results in Table \ref{res1} highlight our model's state-of-the-art performance (3 out of 4) across several sub-protocols. Specifically, our model with HTER dropping to 7.71\%, a 34.44\% decrease from the baseline, and AUC rising to 97.72\%, a 41.27\% increase from the baseline in `PSW$\rightarrow$C' setting. These improvements are a testament to the BIG-MoE architecture's superiority in handling generalized multimodal FAS tasks, indicating the excellent generalization capacity of our model across unseen scenarios.

\vspace{0.5em}
\noindent\textbf{Limited Source Domains Scenario.} \quad   
The results of `PS$\rightarrow$CW' in Table \ref{limited} also demonstrate our model's superior generalization performance under limited source domains, demonstrating enhanced multimodal generalization over `ViT (Baseline)'. With the AUC of 95.13\% and the HTER of 14.11\%, our model leads in state-of-the-art generalization performance, highlighting the model's outstanding ability to generalize across limited source domains scenarios

\vspace{-1.0em}
\subsection{Ablation Study}
\vspace{-0.5em}

\begin{figure}
\centering
\vspace{-2.8em}
\includegraphics[width=0.46\textwidth]{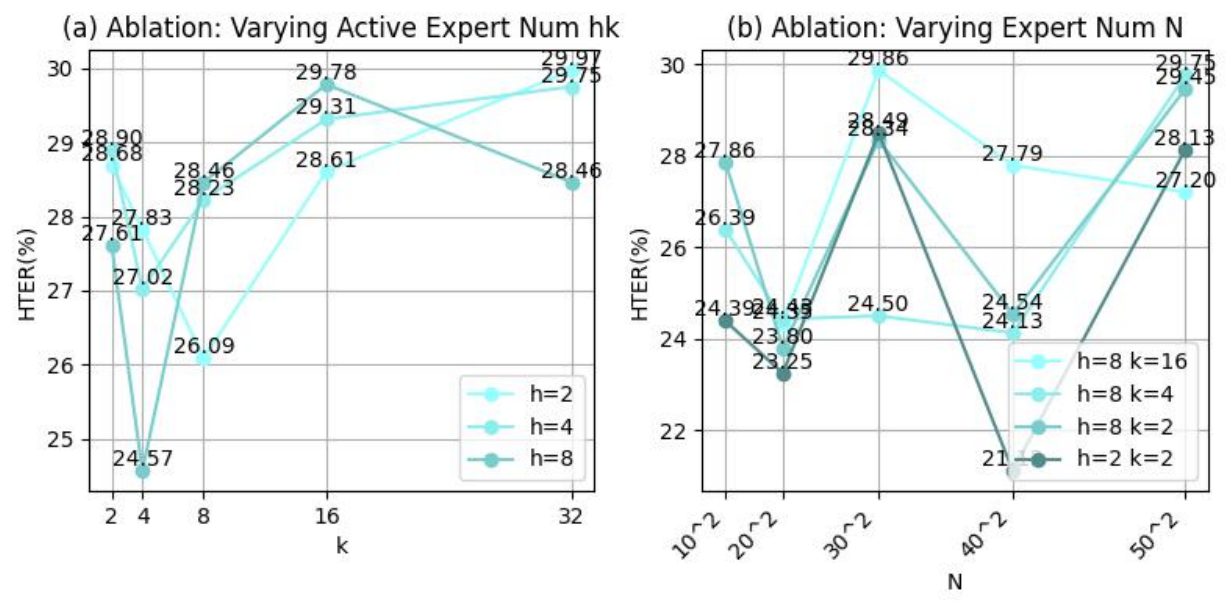}
\vspace{-1.5em}
\caption{Ablation study on expert numbers and activations. (a) HTER with Varying Numbers of Activated Experts. (b) HTER with Different Total Expert Counts. The ablation study investigates the impact of expert count and activation on model performance, providing insights into the optimal configuration for expert utilization in the model.}
\label{Ablation}
\vspace{-1.8em}
\end{figure}

To validate the rationality and effectiveness of BIG-MoE, we conducted meticulous ablation experiments. These aimed to evaluate the impact of prompt settings on model performance, comparing BIG-MoE with a Vision Transformer, a coarse-grained MoE (ST MoE), and a fine-grained MoE (PEER) to highlight the advantages of our CPB and IGMA. Experiments were conducted using the CPW$\rightarrow$S configuration, testing prompt settings with $P_s$ alone, $P_s$ with $P_d$, and the full setup. Results showed that all prompt configurations improved performance, substantiating the rationality and effectiveness of our approach and demonstrating BIG-MoE's potential to enhance model capabilities, providing insights for future work.

\vspace{0.2em}
\noindent\textbf{Impact of Experts' Granularity.} \quad 
Fig. \ref{Ablation} indicates that while moderate increases in IGMA granularity enhance the Adapter's performance, overly fine granularity can lead to a decline in effectiveness. This suggests a critical trade-off: granularity must be judiciously adjusted to maximize system performance, underscoring the need for a balanced granularity strategy in model optimization.

\vspace{0.2em}
\noindent\textbf{Effectiveness of IGMA.} \quad   
The data `w/ IGMA+CPB (With $P_t$) ' in Table \ref{prompt} indicates that, with the help of perceptual cues, IGMA sees a 15.03\% HTER reduction and a 12.94\% AUC increase over the resluts of PEER \cite{he2024mixture}. The integration of fine-grained experts with cues in the IGMA framework maximizes performance, surpassing the benefits of prompts alone. The framework is indispensable for achieving optimal results.

\vspace{0.2em}
\noindent\textbf{Effectiveness of CPB.} \quad   
The results from `w/ IGMA' to `BIG-MoE' in Table \ref{prompt} delineate the significant performance enhancement attributable to each prompt element, thereby validating our design rationale. These findings not only demonstrate the synergistic effects across modalities and features, but also highlight the substantial refinement in cue detection and decision-making capabilities afforded by an optimal prompt combination.

\begin{table}[t]
\centering
\vspace{-2.8em}
\caption{Ablation results on the proposed BIG-MoE. 
} 
\vspace{-0.5em}
\resizebox{0.45\textwidth}{!}{\begin{tabular}{c|cc} 
\hline
  \multirow{2}{*}{\textbf{Method}}&\multicolumn{2}{c}{\textbf{CPW$\rightarrow$ S}}\\
  &HTER(\%)$\downarrow$&AUC(\%)$\uparrow$\\
  \hline
  ViT \cite{dosovitskiy2020image} (Baseline)&20.88 &84.77 \\
  w/ ST MoE \cite{zoph2022st}&14.31&88.69\\
  w/ PEER \cite{he2024mixture}&22.34&84.97\\
  \textbf{w/ IGMA}&21.12 &85.50\\
  \textbf{w/ IGMA+CPB (With $P_t$)}& 20.55&88.41\\
  \textbf{w/ IGMA+CPB (With $P_t\&P_c$)}& 10.44&93.87\\
  \textbf{BIG-MoE (Ours)}&\textbf{10.96} &\textbf{94.35}\\
  
  \hline
\end{tabular}}
\label{prompt}
\end{table}

\begin{figure}[!t]
\centering
\vspace{-0.7em}
\includegraphics[width=0.47\textwidth]{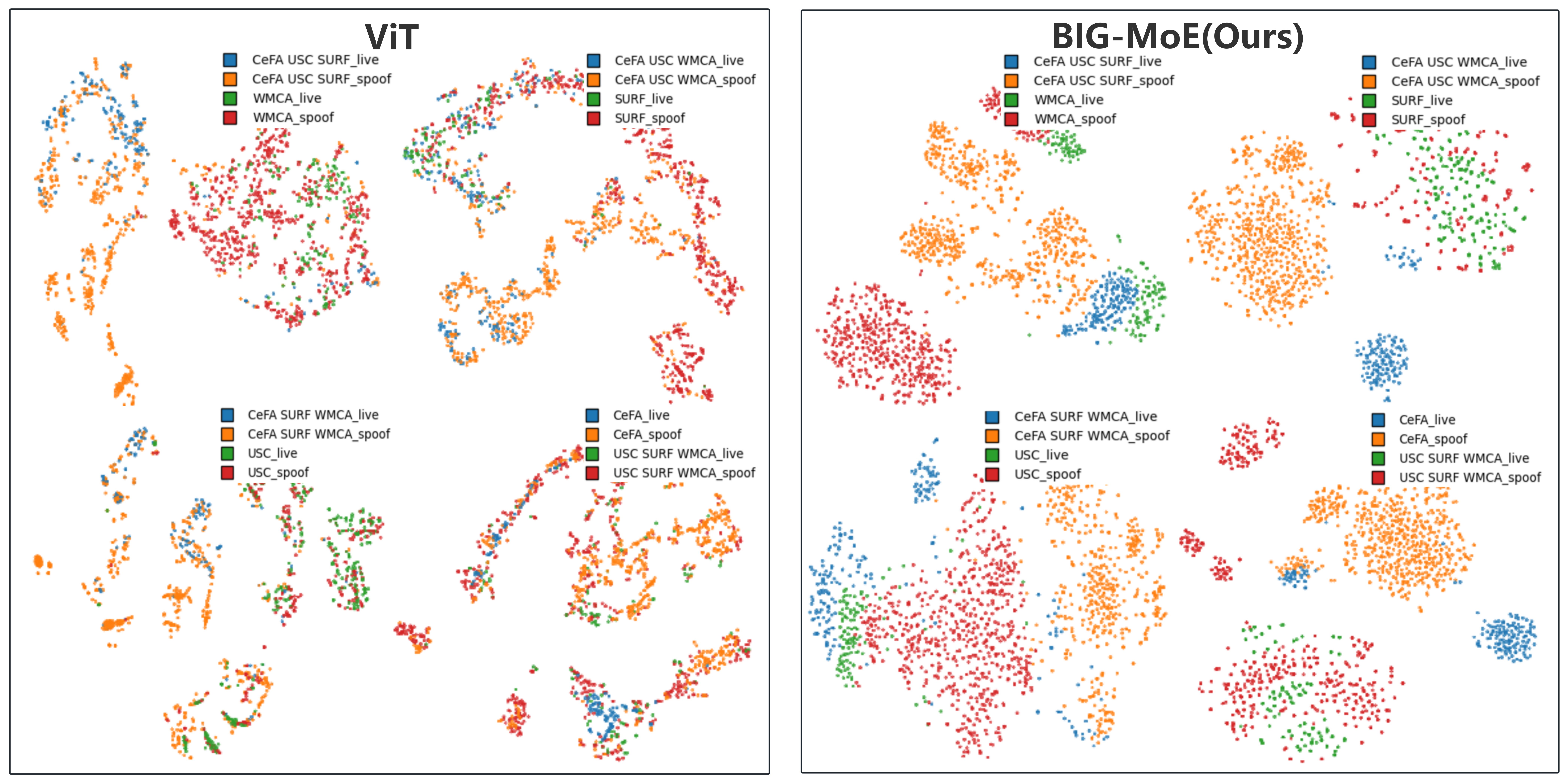}
\vspace{-0.5em}
\caption{t-SNE visualization when respectively tested on CeFA, PADISI, SURF, and WMCA domains.}
\label{tsne}
\vspace{-1.5em}
\end{figure}

\vspace{-0.9em}
\subsection{Visualization and Analysis}
\vspace{-0.5em}
t-SNE was used for dimensionality reduction and visualization of complex data, effectively showing its utility with ViT and BIG-MoE methods. Fig. \ref{tsne} illustrates BIG-MoE's advanced classification, aided by CPB technology in capturing fine feature differences. However, to enhance model generalizability across domains, optimizing multi-domain training samples is needed due to variations in feature representation from different training datasets.

\vspace{-0.6em}
\section{Conclusion}
\vspace{-0.6em}
This paper introduces BIG-MoE, integrating the Isolated Gating Mechanism Adapter and Convolutional Prompt Bypass for generalized multimodal face anti-spoofing (FAS). The former detects subtle spoofing cues with fine-grained experts and efficient key retrieval, while the latter extracts local features and boosts model perception via attention mechanisms. Our method demonstrates superior performance in generalized multimodal FAS through extensive experiments. Future work will focus on improving MoE's generalization with limited samples and in multimodal settings.\\
\noindent\textbf{Acknowledgement.} \quad
This work was supported by National Natural Science Foundation of China (Grant No. 62306061, 82261138629, 62276170), Guangdong Basic and Applied Basic Research Foundation (Grant No. 2023A1515140037, 2023A1515010688), and Open Fund of National Engineering Laboratory for Big Data System Computing Technology (Grant No. SZU-BDSC-OF2024-02), and Guangdong Provincial Key Laboratory under Grant 2023B1212060076. 

\clearpage
\bibliographystyle{IEEEbib}
\bibliography{refs}
\end{document}